\pdfoutput=1
\documentclass[11pt]{article}

\usepackage[final]{acl}

\usepackage{times}
\usepackage{latexsym}

\usepackage[T1]{fontenc}
\usepackage[utf8]{inputenc}

\usepackage{microtype}

\usepackage{inconsolata}

\usepackage{graphicx}

\usepackage{cleveref}
\usepackage{booktabs}
\usepackage{subcaption}
\usepackage{array}

\usepackage{caption}
\usepackage{subcaption}

\usepackage{booktabs}
\usepackage{multirow}
\usepackage{makecell}
\usepackage{arydshln}

\usepackage{microtype}

\usepackage{graphicx}
\usepackage{booktabs}

\usepackage{hyperref}
\usepackage{enumitem}
\usepackage[title]{appendix}
\usepackage{soul}
\usepackage[skins]{tcolorbox}
\usepackage{graphicx}
\usepackage{enumitem}
\usepackage{makecell}

\usepackage{annotation}
\usepackage{tikz}
\usepackage{hyperref}
   
\newcommand{\dataset}{MS-AMI}

\tcbuselibrary{listingsutf8}
\global

\tcbset{
  aibox/.style={
    width=474.18663pt,
    top=10pt,
    colback=white,
    colframe=black,
    colbacktitle=black,
    enhanced,
    center,
    attach boxed title to top left={yshift=-0.1in,xshift=0.15in},
    boxed title style={boxrule=0pt,colframe=white,},
  }
}
\newtcolorbox{AIbox}[2][]{aibox,title=#2,#1}

\title{Tell me what I need to know: Exploring LLM-based (Personalized) Abstractive Multi-Source Meeting Summarization}

\author{Frederic Kirstein\textsuperscript{1,2,*}, Terry Ruas\textsuperscript{1}, Robert Kratel\textsuperscript{1}, Bela Gipp\textsuperscript{1} \\
  \textsuperscript{1}University of Göttingen, Germany \\
\textsuperscript{2}Mercedes-Benz AG, Germany \\
\textsuperscript{*}\texttt{kirstein@gipplab.org} }

\begin{document}
\maketitle

\maketitle
\AddAnnotationRef

\begin{abstract}
Meeting summarization is crucial in digital communication, but existing solutions struggle with salience identification to generate personalized, workable summaries, and context understanding to fully comprehend the meetings' content.
Previous attempts to address these issues by considering related supplementary resources (e.g., presentation slides) alongside transcripts are hindered by models' limited context sizes and handling the additional complexities of the multi-source tasks, such as identifying relevant information in additional files and seamlessly aligning it with the meeting content.
This work explores multi-source meeting summarization considering supplementary materials through a three-stage large language model approach: identifying transcript passages needing additional context, inferring relevant details from supplementary materials and inserting them into the transcript, and generating a summary from this enriched transcript.
Our multi-source approach enhances model understanding, increasing summary relevance by $\sim$9\% and producing more content-rich outputs.
We introduce a personalization protocol that extracts participant characteristics and tailors summaries accordingly, improving informativeness by $\sim$10\%.
This work further provides insights on performance-cost trade-offs across four leading model families, including edge-device capable options.
Our approach can be extended to similar complex generative tasks benefitting from additional resources and personalization, such as dialogue systems and action planning.

\end{abstract}

\section{Introduction}
\label{sec:introduction}
\begin{figure*}
    \centering
    \includegraphics[width=1\linewidth, height=5.5cm]{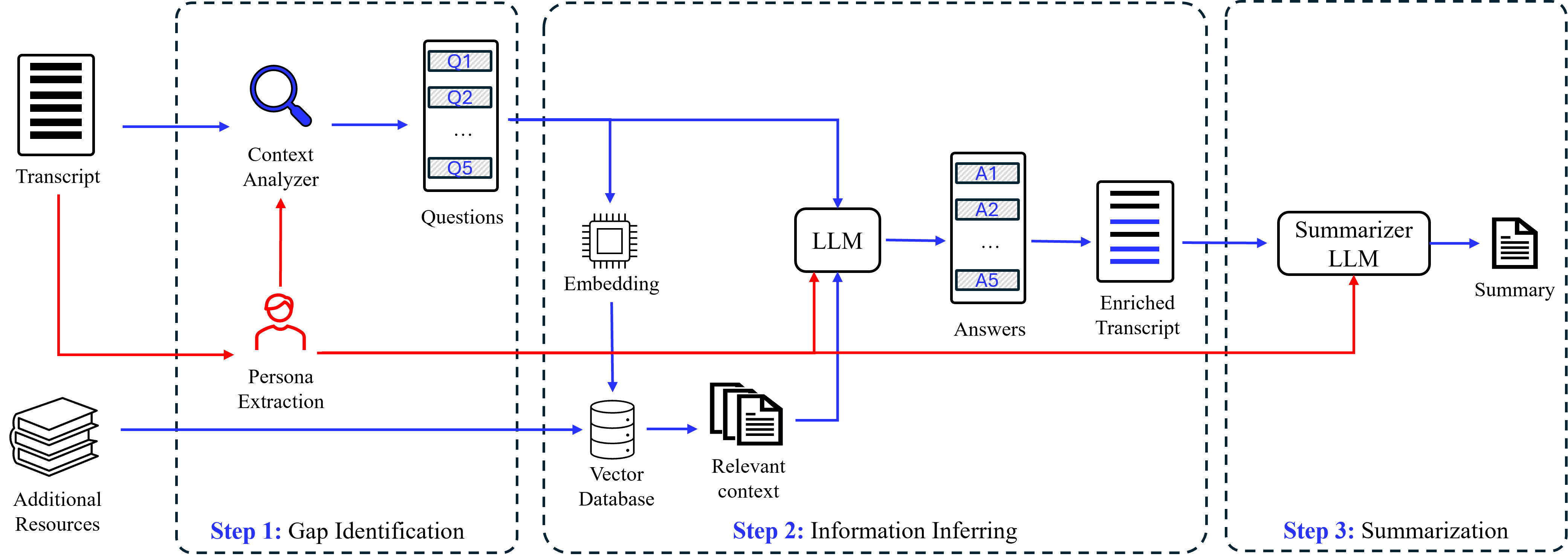}
    \caption{Overview of the three-stage summarization pipeline. Blue boxes and arrows indicate the general pipeline. Red indicates the additional personalization modules.}
    \label{fig:main_figure}
\end{figure*}

Meeting summaries play a key role in professional settings \cite{ZhongYYZ21g, HuGDD23a, LaskarFCB23b}, serving as references, updates for absentees, and reinforcements of key topics discussed.
Major virtual platforms (e.g., Zoom\footnote{\href{https://www.zoom.com/en/ai-assistant}{https://www.zoom.com/en/ai-assistant}}, Microsoft Teams\footnote{\href{https://copilot.cloud.microsoft}{https://copilot.cloud.microsoft}}, Google Meet\footnote{\href{https://support.google.com/meet/answer/14754931}{https://support.google.com/meet/}}) offer summarization systems already, highlighting their importance.
Current methods rely solely on transcripts \cite{ZhuXZH20d,ZhongYYZ21g} and generate generic summaries, often failing to contextualize long discussions' content \cite{KirsteinWGR24a} and to tailor information to individual preferences and productivity requirements.
As such, there is a need for improved model comprehension and personalization in meeting summarization.

% << Stating the problem with multi-source summarization >>
Additional content-related sources can be considered during the summarization process to enhance model comprehension, turning the task into multi-source summarization.
However, traditional approaches of appending documents to transcripts are often limited by model context sizes (e.g., LED \cite{BeltagyPC20d}, DialogLED \cite{ZhongLXZ22g}, Llama \cite{TouvronLIM23a}). 
While hierarchical \cite{ZhuXZH20d} and graph-based methods \cite{PasunuruLBR21a} have been explored, they struggle with handling redundant or contradicting information and maintaining coherence throughout the additional input \cite{MaZGW23}.
Recent advancements in question-answering, which face a conceptually close challenge when answering a query considering an arbitrarily large amount of sources\cite{ChenFWB17}, suggest Retrieval Augmented Generation (RAG) \cite{LewisPPP21a} as a promising solution that efficiently filters relevant information from extensive document collections and uses language models to perform a task such as information inferring.
As RAG is not designed to identify contextual gaps in transcripts, a targeted approach is needed to pinpoint specific information requirements within the transcript, using RAG for focused retrieval.
Otherwise, language models, already challenged by meeting summarization complexities (e.g., omission, repetition, irrelevance) \cite{KirsteinWGR24a}, may become overwhelmed by considering relevant documents in the summarization process.

% << What we propose >>
Our proposed multi-source summarization pipeline (\Cref{fig:main_figure}) mimics a human summarization process and distributes the inherent tasks of multi-source summarization across multiple large language models (LLMs) and an RAG framework.
Our three-stage process, informed by multi-hop question-answering techniques \cite{WangDLW24} and recent research in meeting summarization \cite{KirsteinRG24}, includes: (1) gap identification where an LLM analyzes the transcript, identifies context-deficient passages, and generates questions about missing information; (2) information inferring, using RAG to retrieve relevant documents, process these questions, and insert inferred answers into the transcript; and (3) enriched transcript summarization, where a final LLM generates an abstractive summary.

%%%%%%
%% personalized meeting summarization:
%
Personalized summaries are valuable in professional settings, as participants often write notes focused on points relevant to their projects and knowledge \cite{KhodakeKBH23}.
Current research on personalization mainly explores post-processing techniques \cite{ChenDY23a,JungSJC23}. % or graph-based approaches linking salient turns to speakers depending on X. \cite{}.
We explore salience control and personalization by extending our multi-source meeting summarization pipeline to automatically understand the target audience from the transcript.
Inspired by \citet{GiorgiLAI24}, we incorporate an upstream LLM to extract participant information such as personality traits, project interests and observed knowledge level in the transcript.
These characteristics are used to tailor a summary based on the participant's needs according to the identified gaps.

% adapt after discussion finalized.
We evaluate our pipeline using \dataset{}, a 125-sample multi-source dataset based on AMI \cite{MccowanCKA05}.
Our approach improves informativeness (+0.18 points) and relevance (+0.40 points) compared to single-source summarization, outperforming multi-source summarization with simple document concatenation.
The pipeline shows better contextual understanding and provides more in-depth, relevant information.
Our personalization protocol further enhances informativeness (+0.33 points over a simple personalized baseline) for target readers, tailoring the content to individual preferences.
While using GPT-4 Turbo \footnote{We will refer to this as GPT4 throughout the paper.} \cite{OpenAIAAA24} as our primary model, we also assess smaller models from Phi \cite{AbdinJAA24}, Gemini \cite{GeminiTeamRST24}, and Llama families for practical deployment scenarios.
Overall, GPT4 offers superior performance at the highest cost, while Phi-3 mini provides a cost-effective alternative with similar quality but requires additional robustness measures for personalization.
Our contributions are summarized as follows:
\begin{itemize}[parsep=0.3pt,itemsep=0.1pt]
    \item \dataset, a dataset of 125 meeting summaries and related additional resources. 
    \item A multi-source meeting summarization pipeline that generates and inserts informative comments into meeting transcripts.
    \item Personalization of summaries by embodying participants and their inherent knowledge.
\end{itemize}

The dataset and code ware available through Huggingface and the project-accompanying Github repository:

\begin{center}
\url{https://github.com/FKIRSTE/emnlp2024-personalized-meeting-sum}    
\end{center}

% \section{LLM-based Multi-Source Meeting Summarization}
% \label{sec:MSMS}
% \input{text/03_MultiSourceSum}

% \section{Personalization of MSMS}
% \label{sec:personalized-MSMS}
% \input{text/04_PersonalizedMSS}

\section{Methodology}
\label{sec:methodology}
Our multi-source RAG-based summarization pipeline (\Cref{fig:main_figure}) enriches meeting transcripts with inferred information from supplementary materials, turning the multi-source summarization into a single-source task.
An optional personalization protocol tailors summaries to specific readers by extracting participant information from the transcript and providing this info as the target audience to the generating LLMs.
Leveraging LLMs' zero-shot capabilities \cite{LiuLHP23}, proven effective for meeting summarization \cite{LaskarFCB23b, KirsteinRG24}, our approach is suitable for real-world applications lacking in-domain datasets.
Prompt templates are detailed in \Cref{sec:prompt_templates_appendix}.

\subsection{General Summary Pipeline}
Our multi-source summarization pipeline enhances model comprehension through three stages mimicking the human summarization approach considering additional sources: identifying where additional context is required (\textit{gap identification}), extracting and inferring relevant information from the additional resources (\textit{information inferring}), and summarizing the transcript considering the new information (\textit{summarization}).

In gap identification, an LLM uses chain-of-thought reasoning \cite{WeiWSB23c} to identify and prioritize context-deficient passages, inspired by research on knowledge gap detection in reasoning \cite{WangDLW24} and LLM knowledge \cite{YinSGW23, FengSWD24}.
We further process the identified gaps by having the LLM generate questions about the missing context observed.
A RAG framework then processes these questions, using similarity measures to determine content relevance \cite{LewisPPP21a} and infer answers from relevant sources.
These answers hold the information bits the summarizing system misses to fully comprehend the meeting content and are inserted into the original transcript as comments (see \Cref{sec:example_comment_appendix} for an example).
Finally, an LLM produces an abstractive summary of the enriched transcript \cite{LaskarFCB23b, KirsteinRG24}. 
This approach incorporates supplementary materials, distributing the additional challenges of multi-source summarization (i.e., additional content understanding, salient content extraction, linking to the original transcript) across multiple model instances, without requiring domain-specific training or few-shot examples.

\subsection{Personalized Summary Pipeline}
Meeting summaries are crucial for post-meeting processing and action planning, necessitating personalized, user-centric approaches.
To improve personal efficiency and information retention, the summary should contain what content the reader is most interested in, considering factors such as project relevance or moments of distractions, ideally without the need to manually input constraints \cite{ChenDY23a} or query the transcript \cite{JungSJC23}.
Our personalization protocol leverages an additional LLM to extract target reader details and generate a persona \cite{Paoli23}, i.e., a description regarding personality traits, viewpoints, interests, and additional task-relevant information.
We leverage zero-shot abilities to detect standpoints\cite{LanGJL24}, personalities\cite{RahmanH22,YanMLM24} and knowledge levels \cite{BaekCCH24, CamaraE22}.
An example is shown in \Cref{fig:example_persona} (\Cref{sec:performance_persona}).
The LLM then embodies this persona \cite{SerapioGarciaSCS23,StockliJLH24, LaCavaT24} for gap identification to generate questions from the individual's perspective and informs the RAG and summarizer LLM about their target audience to accordingly tailor the summary.

\section{Dataset}
\label{sec:dataset}
\label{human_annotation}
% FK

For our experiments, due to the lack of an established multi-source meeting summarization dataset, we introduce \dataset{}, an adapted version of AMI \cite{MccowanCKA05}, comprising 125 staged business meetings with processed supplementary content (whiteboard drawings, slides, notes). 
Using GPT-4o\footnote{gpt-4o-2024-05-13} for OCR and image description \cite{ShahriarLMA24}, and Aspose\footnote{aspose-words 24.7.0, Aspose.Slides 24.6.0} for document text extraction, we create a multi-source dataset compatible with language models.
Each meeting's data is compiled into a JSON file, preserving original structures.
We remove 12 samples from the initial 137 meetings due to processing errors.
Dataset statistics are in \Cref{tab:statistics_MSM}, with quality assessment details in \Cref{appendix:dataset_quality}.

% Throughout this work, we leverage an adapted version of established AMI samples \cite{MccowanCKA05}, \dataset{}, that comprises 125 staged business meetings AMI scenario meetings and their original supplementary content (i.e., whiteboard drawings, presentation slides, hand-written notes).
% While previous works overly consider the transcript part of AMI, we process these additional sources to make them compatible with language models and form a multi-source dataset.
% We employ GPT-4o\footnote{gpt-4o-2024-05-13} for OCR to extract text from handwritten notes and generate descriptions of whiteboard drawings \cite{ShahriarLMA24}. 
% Aspose \footnote{aspose-words 24.7.0, Aspose.Slides 24.6.0} is used for text extraction from Word documents and presentations.
% The processed texts are compiled into a single JSON file per meeting, preserving the original structure of supplementary materials.
% We filter the original 138 scenario meetings, removing 13 samples due to erroneous files.
% Comprehensive dataset statistics are provided in \Cref{tab:statistics_MSM}.
% The quality assessment of \dataset{} is discussed in \Cref{appendix:dataset_quality}.

% significant

\section{Experiments}
\label{sec:experiments}
This section explores the quality of summaries generated by our general and personalized pipeline.
We analyze the performance of different LLMs on persona extraction, question generation, and answer generation in \Cref{sec:performance_appendix}.

\paragraph{Setup.}
We use GPT4\footnote{gpt-4-turbo-2024-04-09, default settings, temperature = 0} instances as the backbone model for all stages, leveraging its proven summarization capabilities \cite{LaskarFCB23b, KirsteinRG24}.
Smaller, more practical models are explored in \Cref{sec:practical}.
Throughout the experiments, we set a limit of 250 tokens for the summaries, aligning with comparable works using the AMI corpus or similar datasets \cite{KirsteinRG24}.
% Token limits can also be considered part of personalization, such that it becomes subjective if a summary is too long or too short.
% We leave to future work the development of a personalization approach that plans the optimal length of a summary, extending the work by \citet{WangZWL22}.
We prompt the gap-identifying LLM to point out the top five most relevant context gaps to keep a balance between considering multiple resources and computation effort.
%While dynamically choosing this number appears to be a reasonable approach, we will leave this to future work to devise a method of how the number of gaps can be reliably tied to models’ context understanding and personal behavior through user interaction anticipation.
For RAG, we employ OpenAI's text-embedding-3-small for contextualized embedding \cite{BurganKL24} and cosine similarity for distance measurement. 
Large documents are chunked to fit into the embedding model.

\paragraph{Evaluation.}
\label{sec:refinement-evaluation}
For evaluation, we use AUTOCALIBRATE \cite{LiuYHZ23a} and a GPT4-powered metric assessing following the theoretical concept of FACTSCORE \cite{MinKLL23} (i.e., breaking sentences down into atomic facts which are compared to the transcript regarding factuality) to report 5-point Likert score to assess content coverage, salience, and overall quality in the categories: relevance (REL), informativeness (INF), factuality (FAC) and overall (OVR):
\begin{itemize}
    \item Informativeness (INF): Assesses completeness and clarity. Ensures all essential details and key ideas are conveyed without omissions or ambiguity.
    \item Relevance (REL): Measures alignment with (user's) specific information needs. Focuses on inclusion of central key points.
    \item  Factuality (FAC): Refers to accuracy and truthfulness. Ensures all information is consistent with the original content.
    \item Overall (OVR): Assesses the overall summary quality using error types defined by \cite{KirsteinRG24}.
    These include redundancy, incoherence, language issues (i.e., inappropriate or ungrammatical usage, and failure to capture unique styles), omissions, coreference problems (i.e., unresolved references, misattributions, or missing mentions), hallucinations, structural flaws (i.e., misrepresenting discourse order or logic), and irrelevance. 
    The generated Likert-score (1-5) reflects the summary's performance across all these categories, providing a comprehensive evaluation of its quality and accuracy.
\end{itemize}
We extend the same metrics for personalized summaries considering extracted personas (category-P).
We assess the matching with a set of human-generated labels, achieving accuracies of REL: 87.3\%, INF: 92.4\%, FAC: 85.7\%, OVR: 93.6\%, REL-P: 91.5\%, INF-P: 89.8\%, and OVR-P: 87.8\%
Further details on the evaluation are stated in \Cref{appendix:evaluation_details}.

\subsection{Results and discussion on the general multi-source summarization pipeline}
\label{sec:general_pipeline}

\begin{table}[]
    \centering
    \small
    \begin{tabular}{lccccccc}
    \toprule
    \textbf{Setup}       & \textbf{INF}   & \textbf{REL}   & \textbf{FAC} & \textbf{OVR}\\
    \midrule
    G-infer     & \textbf{4.49*} & 4.04** & \textbf{4.78*} & \textbf{4.41*} \\
    G-top & 4.33 & 4.02** & 4.67* & 4.30\\
    G-all & 4.40 & \textbf{4.11**}  & 4.30 & 4.35* \\
    \midrule
    G-none   & 4.31 & 3.70 & 4.33 & 3.99 \\
    GOLD        & 3.79 & 3.59 & 4.98* & 4.12 \\
    \bottomrule
    \end{tabular}
    \caption{LLM-based 5-point Likert scoring of the general multi-source meeting summarization pipeline. Significant values: * (p $\leq$ 0.05) and ** (p $\leq$ 0.01). Best scores are \textbf{bold}.}
    \label{tab:general_pipeline_results}
\end{table}

\paragraph{Baseline.}
We compare our multi-source pipeline (G-infer) against three baselines.
G-none is a single GPT4 model without access to additional information.
G-all is given all available additional sources appended to the transcript's end.
G-top considers only the top 5 closest additional sources based on an RAG framework.
GOLD refers to the huamn generated summary.

\begin{table}[]
\centering
\small
\begin{tabular}{lcccc}
\toprule
\textbf{Setup}       & \textbf{INF-P}   & \textbf{REL-P}   & \textbf{FAC}   & \textbf{OVR-P} \\
\midrule
P-infer+per & \textbf{4.51*} & 4.16* & 4.65* & \textbf{4.79*} \\
P-per & 4.43* & \textbf{4.18*} & 4.59* & 4.50* \\
\midrule
P-infer & 4.34 & 4.09 & \textbf{4.75*} & 4.35 \\
P-all & 4.18 & 4.04 & 4.38 & 4.20 \\
\midrule
P-none & 4.00 & 3.59 & 4.33 & 4.03 \\
\bottomrule
\end{tabular}
\caption{LLM-based 5-point Likert scoring of the personalized multi-source meeting summarization pipeline. Best scores are \textbf{bold}. Significant values: * (p $\leq$ 0.05) and ** (p $\leq$ 0.01).}
\label{tab:personal_pipeline_results}
\end{table}

\paragraph{Structured inclusion of inferred details enhances multi-source summarization quality.}
\label{sec:personal_pipeline}

Results in \Cref{tab:general_pipeline_results} show that multi-source summarization enhances OVR summary quality by up to 0.42 (G-infer) compared to a single-source model, supporting the general effectiveness of multi-source summarization in general.
Multi-source in general improves REL by at least 0.32 points over G-none.
We derive from this that the structured inclusion of inferred details in the transcript enhances context understanding, clarifies information relationships, and strengthens the summary structure, which is backed by the evaluating LLM's CoT explanation.
G-infer further reduces hallucination, increasing FAC scores by 0.45 over G-none, aligning with recent findings \cite{DasGGR24}.
This improvement likely stems from the model's enhanced ability to ground summaries in concrete, relevant information from multiple sources \cite{LiYZ24}.
INF shows a modest increase (+0.18 from G-none to G-infer), as additional information primarily aids contextualization rather than content representation. % \cite{missing}.
Comparing the different general summary pipelines (G-infer, G-top, G-all), G-infer’s improvements in INF, FAC, and OVR are significant (p $\leq$ 0.05). The relevance score of G-all (4.11) is not a significant improvement over the scores of G-infer (4.04) or G-top (4.02), but all are significantly better than G-None (p $\leq$ 0.01).
This significance underscores that multi-source, in general, improves REL.

Our qualitative analysis (see examples in \Cref{sec:general_sum_examples}) supports these quantitative findings, revealing that multi-source summarization significantly enhances models' transcript contextualization and explanation capabilities.
G-top summaries exhibit the most hallucination and limited context understanding of the multi-source setups.
G-all summaries are prone to repetition errors due to repeated statements in several supplementary files.
G-infer demonstrates the best content understanding and higher content density, though it occasionally includes excessive detail.

Our findings suggest that G-infer is most effective for multi-source summarization, outperforming simple concatenation of all data. 
Concatenating only the top five related sources performs worst of the multi-source approaches, likely due to insufficient information in some documents.
This suggests that selective, context-aware integration of supplementary information is more beneficial than limited or unstructured inclusion.
Alternative similarity measures for RAG beyond cosine similarity \cite{BehnamGhaderMR23, Ampazis24} might improve performance for G-top.

\subsection{Results and discussion on the personalized multi-source summarization pipeline}
We follow the same setup as for the general pipeline, using GPT4 as the backbone model.
Here we add the persona extraction stage to inform the subsequent stages about the participants' traits.

\paragraph{Baseline.}
In addition to our full pipeline (P-infer+per) with RAG-based information insertion and persona consideration, we evaluate the infer, all, and none variants as additional baselines, named P-infer, P-all, and P-none.
Additionally, we consider P-per, where a persona is extracted and provided to the summarization model, but without using the RAG stage.
All variations are informed about the target participant.
We exclude the previously tested G-top variation due to its weaker performance.

\paragraph{Detailed persona inclusion improves personalization but complicates content handling.}

\Cref{tab:personal_pipeline_results} shows that including detailed personas improves INF-P (up to 0.25) and REL-P (up to 0.14) from P-all to P-per, aligning with recent prompt engineering findings \cite{Lovlund24}.
P-per outperforms P-infer in the OVR-P score, indicating the positive influence of the persona consideration when focusing the evaluation on personalization.
Scores vary across participant roles ('Project Manager,' 'User Interface,' 'Marketing,' 'Industrial Design'; see extended results in \Cref{sec:extended_result_tables}), with 'Project Manager' often yielding higher scores, suggesting more insightful persona extraction for some roles.
Deviations up to 0.40 are observed across roles (e.g., P-per pipeline on INF-P).
% Analysis in \Cref{sec:performance_persona} shows no clear tendency for better extraction of specific personas, indicating need for further exploration.
The P-per and P-infer+per REL scores are significant (p $\leq$ 0.05) over the other scores, highlighting the benefit of the persona extraction approach.

Qualitative analysis (examples in \Cref{tab:general_pipeline_examples,tab:personal_pipeline_examples} in \Cref{sec:personal_sum_examples}) shows that persona-based summaries vary significantly in quality, while target-informed pipelines produce more consistent results.
Evaluating LLMs' CoT reasoning reveals that P-all and P-infer pipelines tend to omit content due to a limited understanding of the target audience.
P-infer+per generates the most tailored and relevant summaries, though P-infer slightly outperforms in context understanding and exhibits fewer informativeness-related errors.

We conclude that personalization benefits from extensive reader information, but linking salient content to specific personas remains complex.
This calls for advanced techniques balancing personalization with effective multi-source content integration, potentially using sophisticated algorithms for salience determination and persona-content matching.
A possible improvement could involve an additional critique model \cite{KirsteinRG24} to check generated summaries for features like omission, hallucination, or structure, and propose corrections accordingly.

\subsection{Practical Application}
\label{sec:practical}
\begin{table}[]
    \centering
    \small
    \begin{tabular}{lcccc}
    \toprule
      & \textbf{G-GPT4}    & \textbf{G-Phi3}& \textbf{G-Gem}   & \textbf{G-Llama3}  \\
    \midrule
    INF     & 4.59      & 4.18  & 4.36  & 3.84\\
    REL     & 4.09      & 3.97  & 4.12  & 3.75\\
    FAC     & 4.88      & 4.38  & 4.64  & 4.69\\
    OVR     & 4.34      & 4.12  & 4.24  & 4.06\\
    \midrule
    Cost   & \$0.25  & \$0.007  & \$0.009  & \$0.001\\
    \midrule
    Time   &  110s & 32s  & 92s  & 68s\\
    \bottomrule
    \end{tabular}
    \caption{LLM-based 5-point Likert scores of the general summarization pipeline, comparing different model families. Costs are per sample.}
    \label{tab:budget_pipeline_results_general}
\end{table}
\begin{table}[]
    \centering
    \small
    \begin{tabular}{lcccc}
    \toprule
            & \textbf{P-GPT4} & \textbf{P-Phi3} & \textbf{P-Gem}   & \textbf{P-Llama3}  \\
    \midrule
    INF-P     & 4.44  & 3.97  & 4.54  & 3.85\\
    REL-P     & 4.12  & 3.79  & 4.00  & 3.82\\
    FAC     & 4.48  & 4.40  & 4.43  & 4.35\\
    OVR-P     & 4.54  & 4.36  & 4.49  & 4.00\\
    \midrule
    Cost   & \$0.37  & \$0.01  & \$0.013  & \$0.002\\
    \midrule
    Time (s)   &  152s & 44s  & 114s  & 76s\\
    \bottomrule
    \end{tabular}
    \caption{LLM-based 5-point Likert scores of the personalized summarization pipeline, comparing different model families. Costs are per sample.}
    \label{tab:budget_pipeline_results_personal}
\end{table}

After exploring multi-source summarization and personalization with GPT4, we investigate smaller, more efficient LLMs to assess the practical use of our concepts.
We now evaluate our best pipeline setups ('infer' and 'infer+per') using Phi-3 mini 3.8b 128k (Phi3) \cite{AbdinJAA24}, Gemini Flash 1.5 (Gemini) \cite{GeminiTeamRST24}, and Llama 3 8b\footnote{\href{https://llama.meta.com/llama3/}{https://llama.meta.com/llama3/}} (Llama3) on one-third of \dataset{}\footnote{Models accessed via APIs: GPT4, Phi3 - Azure, Gemini - Google Cloud, Llama3 - Groq.}.
For Llama3, which cannot fit most meetings into its 8k token limit, we employ a sequential chunking approach \cite{ChangLGI24a}.
Examples of generated summaries are shown in \Cref{sec:practical_sum_examples}.

\Cref{tab:budget_pipeline_results_general,tab:budget_pipeline_results_personal} show the performance of the different LLMs used as backbone models for general and personalized summarization.
Results show all models can run our multi-source pipeline, with larger models scoring higher.
Surprisingly, Phi3 often outperforms the larger Llama3, likely due to Llama3's hierarchical summarization limitations.
Qualitatively, Gemini produces high-level but shallow summaries, Phi3 closely matches GPT4 with occasional detail gaps, and Llama3 struggles with repetition and structure.
Notably, Phi3 inconsistently identifies five context gaps, potentially indicating context understanding weaknesses \cite{KirsteinWRG24a}.

Extracted personas are similar across models.
Phi3 and GPT4 produce similar-length personas, while Gemini and Llama3 generate longer, slightly more lengthy ones.
Phi3 often focuses on participant actions rather than reader-relevant information in the generated summaries, suggesting a need for further adaptation.
Llama3 includes irrelevant content, reflected in low INF-P and REL-P scores (\Cref{tab:budget_pipeline_results_personal}).
Gemini handles personalization well but tends towards high-level summaries again, sometimes omitting crucial details.

GPT4 is the most expensive model, Gemini and Phi3 cost similarly, and Llama3 is the cheapest.
Llama3 and Phi3 can also run locally.
Phi3's design for weaker hardware enables further cost reduction and offline on-device usage.
GPT4 takes the longest with $\sim$152 seconds per personalized summary, Phi3 is the fastest by far with $\sim$44 seconds, while Gemini and Llama3, the latter due to additional calls for large inputs, are in between.
General summaries are up to 30\% quicker.

Considering performance, cost, and time, GPT4 excels in unrestricted scenarios, while Phi3 is ideal for constrained environments, offering good-quality on-device summaries but requiring additional quality assurance for personalization \cite{KirsteinRG24}.
Gemini performs similarly to GPT4 with a slight price advantage but less detailed summaries.
Llama3 needs further adaption, likely regarding the hierarchical summarization.

%\section{Scenarios}
%\label{sec:ablation}
%\input{text/05_Real_World_Scenario}

\section{Related Work}
\label{sec:related_work}
% \noindent \textbf{Meeting summarization} solutions are shifting from encoder-decoder models to Large Language Models (LLMs).
% While earlier approaches using BART \cite{LewisLGG20b} and PEGASUS \cite{ZhangZSL20a} are trained to handle specific meeting-related challenges like speaker relations \cite{HuaDM23a} and roles \cite{AsiWEG22a,NarakiSH22a}, recent LLM-based methods achieve comparable performance without domain-specific fine-tuning \cite{LaskarFCB23,KirsteinWRG24}.
% This is due to LLMs' zero-shot capabilities \cite{missing} and improved context understanding, which makes them promising for real-world applications.
% Our approach follows this recent trend and uses GPT4 \cite{EMNLP} as the backbone model, enabling the exploration of new concepts without extensive training data.

\noindent \textbf{Personalized summarization.}
A recent consideration when producing high-quality summaries is related to the identification of saliency for the reader \cite{KirsteinWGR24a}, introduced as personalized meeting summarization by \citet{KhuranaKKS23}, which aims to identify reader-specific salient information.
Unlike existing approaches leveraging graph-based \cite{JungSJC23} or human-in-the-loop methods \cite{ChenDY23a}, we use personas \cite{Paoli23} to guide LLM generation. 
Extending recent works \cite{LanGJL24,YanMLM24}, we extract personality traits, stances, and knowledge from transcripts to steer the detection of context gaps and inform RAG and summarization modules about the target audience.

% https://arxiv.org/pdf/2306.09604
% https://arxiv.org/abs/2403.12968

\noindent \textbf{Multi-source summarization.}
Considering additional resources for summarization is underexplored due to complexities in processing large text spans with traditional architectures \cite{MaZGW23}.
Existing methods like sentence clustering \cite{NayeemFC18a}, graph-based modeling \cite{PasunuruLBR21a}, and hierarchical summarization \cite{ZhuXZH20d} struggle with context understanding \cite{AmplayoL21a} and handling contradicting or redundant content \cite{MaZGW23}.
Inspired by the conventionally close open-domain QA \cite{ChenFWB17}, we explore and leverage RAG to multi-source summarization, identifying context gaps \cite{WangDLW24} and using them for RAG-based answering \cite{LewisPPP21a}.
Our approach uniquely applies these concepts to meeting summarization tasks.

\section{Final Considerations}
\label{sec:final_considerations}
This paper presents a three-step RAG-based pipeline using multiple LLM instances to abstractly summarize Englisch business meeting transcripts, considering supplementary files.
We also explored how to use personas extracted from transcripts to introduce personalization and preferences in summaries.
Key findings show incorporating supplementary sources improves summary quality by at least 0.31 over the baseline (single-source), with an additional 0.11 improvement when distributing multi-source challenges (identifying, inferring, and linking salient content) across multiple sources.
Persona-based personalization, using dynamically generated participant personas, enhances relevance by up to 0.44 compared to a baseline with only the target audience's role information.
Our zero-shot setup performs well with significantly smaller models than GPT-4 turbo, revealing that Phi-3 mini 128k produces good-quality summaries under a low-resource environment.
This study provides initial insights into multi-source and personalized meeting summarization using LLMs and RAG systems, leaving the development of more sophisticated approaches, such as multi-agent discussions for retrieval and personalization, and the development of a dynamic function to identify the best amount of resources to consider to future work.

\section*{Acknowledgements}
This work was supported by the Lower Saxony Ministry of Science and Culture and the VW Foundation.

\section*{Limitations}
Although our proposed MultiSourceMeeting might seem small (125 samples), its size is comparable to the original AMI dataset (137 samples).
We contribute to extending the original datasets with careful alignment and curation of additional resources where available.
Another possible limitation in our work is the use of only GPT4 in our main experiments.
We chose GPT4 because of its large context size (e.g., 128k tokens) and better initial robustness when exploring new concepts.
Another potential drawback is that our pipeline faces challenges in jointly optimizing prompts across different model families, potentially leading to performance variations.
We address this by adapting best practices for individual stage-informing methods and model-specific prompting techniques, translating methodological concepts to fit each backbone model prompt-wise.
Pre-testing was conducted for each stage and model to refine prompts and mitigate obvious limitations before experiments.

\section*{Ethics Statement}
\paragraph{Licenses:}
We adhered to licensing requirements for all tools used (OpenAI, Microsoft, Google, Meta, Huggingface).

\paragraph{Privacy:}
User privacy was protected by screening the dataset for personally identifiable information during quality assessment (\Cref{appendix:dataset_quality}).

\paragraph{Intended Use:}
Our pipelines are intended for business organizations to generate quick, personalized meeting overviews.
While poor summary quality may affect user experience, it should not raise ethical concerns as summaries are based solely on given transcripts.
Production LLMs will only perform inference, not re-training on live transcripts.
Summaries will be accessible only to meeting participants, ensuring information from other meetings remains confidential.

% Bibliography entries for the entire Anthology, followed by custom entries
%\bibliography{anthology,custom}
% Custom bibliography entries only
\bibliography{custom, 24_EMNLP_MultiSourceDataset, emnlp_industry_addendum}

\appendix
\label{sec:appendix}
\section{Dataset Quality Assessment}
\label{appendix:dataset_quality}
%\paragraph{Quality assessment.}
To ensure \dataset{}'s integrity and usability, we conduct a quality assessment using three graduate students \footnote{The origin of the funds and annotators will be disclosed later to avoid the risk of giving the author's identity. The students were paid via their internship.} with diverse academic backgrounds (e.g., computer science, psychology, communication science), English proficiency, and familiarity with meeting summarization.
Each sample undergoes a dual-annotator review focusing on OCR quality, Aspose text extraction, and privacy concerns to assess the quality and perform corrections if necessary.
For OCR, the annotators are asked to look for artifacts changing individual words, and if the generated image descriptions match the drawings.
For Aspose, they assign a label according to if all text is extracted successfully and the original structure maintained.
Regarding privacy concerns, the annotators check all sources to see if any personal information of participants is disclosed that should not be part of the dataset, marking instances.
% Annotators check for word-altering artifacts, image description accuracy, extraction completeness, and inappropriate personal information disclosure. 
In cases of disagreement, a third annotator is consulted.
The assessment reveals consistently accurate GPT-4o extractions, correct alignment across samples, and no privacy risks. 
This comprehensive evaluation process ensures \dataset{}'s reliability and ethical compliance for multi-source meeting summarization research.

Statistics on \dataset{} are listed in \Cref{tab:statistics_MSM}

% \begin{table*}
%   \small
%   \centering
%     \begin{tabular}{lccccccc}
%     \toprule
%     Characteristics & AMI & MB-Seattle & MB-XY & MB-XY & MB-XY & MB-XY & MSM (total)\\
%     \midrule
%     \# Meetings & 125?? & & & & & &\\
%     \# Turns & & & & & & &\\
%     \# Speakers & & & & & & &\\
%     \# Len. of Meet. & & & & & & &\\
%     \# Len. of Gold Sum. & & & & & & &\\
%     \# Len. of Aut. Sum. & & & & & & &\\
%     \# Add. Documents & & & & & & & \\
%     Type Add. Documents & & & & & & &\\
%     \bottomrule
%     \end{tabular}%
%   \caption{EXAMPLE - Sth like this.}
%   \label{tab:statistics_MSM}%
% \end{table*}%

\begin{table*}[]
    \small
    \centering
    \begin{tabular}{ccccccc}
        \toprule
        \# Meetings & \# Turns  & \# Speakers   & Len. of Meet. & Len. of Mod. Meet.    & Len. of Gold  & \# Documents\\
        \midrule
        125         &  558.4    & 4.0           &  6567.9     &  6936.6    & 185.5          & 21.8         \\
        \bottomrule
    \end{tabular}
    \caption{Statistics for the \dataset{}. Values are averages of the respective categories. Lengths (Len.) are in number of words.}
    \label{tab:statistics_MSM}%
\end{table*}

\section{Example of Comment in Transcript}
\label{sec:example_comment_appendix}
The questions pointing out gaps in context are answered from supplementary files, inferring the required information. 
This information is injected into the original transcript as a comment enclosed in \texttt{[]} and placed after passages requiring additional context. 
\Cref{fig:example_comment} provides an illustrative example of this format.

\begin{figure*}[t]
    \begin{AIbox}{Example for comment injection in transcript}
    \parbox[t]{\textwidth}{
        \textbf{Project Manager}\newline
        ... \newline
        So, like, I wonder if we might add something new to the to the remote control market, such as the lighting in your house, or um ...  \newline
        \texttt{[The additional functionalities being considered for the new remote control to enhance its appeal and usability include the ability to control multiple devices, potentially integrating control of household lighting, adding a feature to help locate the remote when it's lost (such as making a noise when a high-pitched sound is made), and possibly incorporating a touchscreen. The design aims to combine as many uses as possible, similar to how palm pilots evolved to include multiple functions like cameras, MP3 players, and telephones. ]} \newline
        Yeah, yeah. An Yeah. Like, p personally for me, at home I've I've combined the um the audio video of my television set and my DVD player and my CD player. \newline
        ...
    }
    \end{AIbox}
    \caption{Example for inferred information injected as comment in squared brackets. }
    \label{fig:example_comment}
\end{figure*}

\section{Evaluation Details}
\label{appendix:evaluation_details}
We use AUTOCALIBRATE \cite{LiuYHZ23a} and GPT4 prompted to follow the concept of FACTEVAL \cite{MinKLL23} for evaluation.
This choice is motivated by its scalability, as human evaluation of over 3000 summaries is infeasible, and because the LLM-based metrics do not require reference summaries, making evaluation of the personalization scenario easier.
ROUGE \cite{Lin04b} and BERTScore \cite{ZhangKWW20b} are not reported as main metrics as they yield nearly identical scores across pipeline variants, limiting their interpretability.
We validate metrics against human judgment\footnote{The guidelines and definitions for the individual metrics will be shared later in the project accompanying GitHub repository.}
 employing the three annotators from \Cref{human_annotation}, having all three rate automatically generated samples using the original AMI gold summaries for general summary samples and personalized summaries generated by GPT-4.
Compared to these human labels, the LLM-based metrics achieve high accuracy (REL: 87.3\%, INF: 92.4\%, FAC: 85.7\%, OVR: 93.6\%). For personalized summaries, extended metrics show accuracy scores of 91.5\% (REL-P), 89.8\% (INF-P), and 87.8\% (OVR-P) accuracy.
Inter-annotator agreement (Krippendorff's alpha \cite{Krippendorff70}) is detailed in \Cref{tab:evaluation_agreement}.

\begin{table}[]
    \centering
    \begin{tabular}{lc}
        \toprule
         \textbf{Metric} & \textbf{Krippendorff's $\alpha$} \\
         \midrule
         INF & 0.834 \\
         REL & 0.813 \\
         FAC & 0.856 \\
         OVR & 0.850 \\
         \midrule
         INF-P & 0.799 \\
         REL-P & 0.854 \\
         OVR-P & 0.813 \\
         \bottomrule
    \end{tabular}
    \caption{Inter-annotator agreement scores for human annotating the different evaluation metrics on the \dataset{} dataset.}
    \label{tab:evaluation_agreement}
\end{table}

\section{Performance of the Pipeline's Subcomponents}
\label{sec:performance_appendix}
\subsection{Persona Extraction}
\label{sec:performance_persona}

Our persona extraction process builds on existing approaches for retrieving standpoints \cite{LanGJL24}, personalities \cite{RahmanH22,YanMLM24}, and knowledge levels \cite{BaekCCH24, CamaraE22}.
We validate extraction accuracy through human assessment, with three annotators (\Cref{human_annotation}) evaluating 120 personas (30 per participant role, i.e., User Experience (UE), Project Manager (PM), Industrial Design (ID), Marketing (M)) for fit with the transcript.
Inter-annotator agreement scores are in \Cref{tab:persona_agreement}.
Results show GPT4 reliably extracts personas (acceptance rates: UE 83\%, PM 94\%, ID 87\%, M 92\%), with generated personas differing among participants and slightly across meetings, reflecting evolving standings, knowledge, and interests.
An example persona is shown in \Cref{fig:example_persona}.

% We build our persona extraction process on a set of existing approaches to retrieve standpoints \cite{LanGJL24}, personalities\cite{RahmanH22,YanMLM24}, and knowledge levels \cite{BaekCCH24, CamaraE22}.
% As we ask the backbone model to retrieve all these details at once, we perform a small human assessment to confirm that the extraction is correct.
% We, therefore, ask our three annotators from \Cref{human_annotation} to assign a given persona and, according to the transcript, a binary yes/no according to the question 'Does the described persona fit the participant?'.
% The annotators assess 120 personas in total, 30 for each participant (User Experience (UE), Project Manager (PM), Industrial Design (ID), Marketing (M)).
% The inter-annotator agreement scores are stated in \Cref{tab:persona_agreement}.
% Overall, the human assessment reveals that GPT4 is capable of extracting personas reliably (acceptance: UE: 83\%, PM: 94\%, ID: 87\$, M: 92\%) and that the generated personas are different among the participants and slightly differ between meetings for the same participants, reflecting the developing standings, knowledge, and project related interests.

\begin{table}[]
    \centering
    \begin{tabular}{lc}
        \toprule
         \textbf{Role} & \textbf{Krippendorff's $\alpha$} \\
         \midrule
         UE & 0.734 \\
         PM & 0.796 \\
         ID & 0.753 \\
         M  & 0.728 \\
         \bottomrule
    \end{tabular}
    \caption{Inter-annotator agreement scores on assessing if the extracted personas match participants on samples of the \dataset{} dataset.}
    \label{tab:persona_agreement}
\end{table}

\begin{figure*}[t]
    \begin{AIbox}{Example Persona}
    \parbox[t]{\textwidth}{
    \textbf{Role:} \newline 
    The participant, Laura, is the Project Manager for the project of designing a new remote control. Her role involves organizing the meeting, setting the agenda, ensuring that the project's objectives are clear, and guiding the team through the design process. \newline
    \textbf{Personality Traits (Big Five):} \newline
    \begin{itemize}[parsep=0.3pt,itemsep=0.1pt]
        \item Openness: Laura demonstrates openness through her willingness to explore new ideas for the remote control, such as considering additional functionalities and design commitments beyond traditional features.
        \item Conscientiousness: As a Project Manager, Laura is organized, punctual (noting that everyone arrived on time), and focused on the project's financial constraints and design stages.
        \item Extraversion: Laura's extraversion is not explicitly shown in the transcript, but she seems comfortable leading the meeting and engaging with the team.
        \item Agreeableness: She shows agreeableness by encouraging participation, valuing team members' input, and being considerate of their contributions.
        \item Neuroticism: There is no clear evidence of high neuroticism; Laura appears to handle the meeting's flow and potential stressors professionally.
    \end{itemize}
    \textbf{Point of Views:} \newline
    Laura values user-friendly, original, and trendy design for the new remote control.She is aware of the importance of international market considerations and the need for the product to appeal to different regions. Laura recognizes the need for the remote control to have a unique selling point and possibly additional features beyond controlling the television. \newline
    \textbf{Contributions:} \newline
    Laura facilitates the meeting, introduces the project's brief, and outlines the design stages. She prompts discussions on the product's international appeal, cost considerations, and potential features. Laura ensures that the team stays on track with the meeting's agenda and time constraints. \newline
    \textbf{Knowledge Brought to the Meeting:} \newline
    Laura is knowledgeable about the project's objectives, the design brief, and the financial targets (e.g., selling price and production cost). She understands the importance of the product being original, trendy, and user-friendly. Laura is aware of the need to consider international markets and the potential for different functionalities in the remote control. \newline
    \textbf{Information Not Known:} \newline 
    Laura is unsure about the specifics of what the team members received in their emails, indicating a potential gap in communication or information dissemination. She is uncertain about whether the selling price is wholesale or retail, showing a need for clarification on pricing strategy. Laura seeks input from the team on various aspects, such as design preferences and market considerations, indicating that she does not have all the answers and relies on the team's expertise. \newline
    \textbf{Other Relevant Information:}\newline 
    Laura is attentive to the team's dynamics, ensuring everyone is introduced and feels included in the discussion. She is practical, focusing on the project's constraints and the need for the design to be financially viable. Laura is open to innovation, as seen in her interest in adding new features to the remote control and considering different design aspects like streamlining and touchscreens.\newline
    \newline
    In summary, Laura, as the Project Manager, is an organized and open-minded leader who values collaboration, innovation, and practicality in the design process. She is focused on meeting the project's objectives while being receptive to the team's expertise and input.
    }
    \end{AIbox}
    \caption{Example of an extracted persona for 'Project Manager'.}
    \label{fig:example_persona}
\end{figure*}

\subsection{Gap Identification and Question Generation}
\label{sec:performance_gap_question}
Our gap identification approach builds on work identifying gaps in LLM knowledge \cite{FengSWD24,YinSGW23} and in texts forming the base to answer reasoning tasks \cite{WangDLW24}.
We find that the capabilities of the language models used there (e.g., Llama 2, GPT-3.5) also transfer to GPT4, which successfully generates questions on contextual gaps not directly covered in the transcript.
These questions are often strategic (e.g., "Has the team considered the implications of using speech recognition technology, and what are the arguments for and against its inclusion?"), providing a global perspective and enhancing contextualization.
For personalization, questions vary based on the persona embodied by the LLM, such as User Experience ("What are the implications of omitting the numeric keypad in terms of user navigation and channel selection efficiency?") versus Marketing ("Can we clarify the specific consumer preferences regarding the importance of appearance over functionality for our remote control design?").
This indicates successful persona embodiment and viewpoint-specific questioning, adapting to different roles and perspectives within the meeting context.

\subsection{Information Inferring and Answer Generation}
\label{sec:performance_inferring_answer}
Answering questions based on a set of retrieved, related works, follows the core concept of RAG \cite{LewisPPP21a}.
GPT4 performs well as generating model in such a setup \cite{HoOCZ24}, and also reliably answers questions in our pipeline using RAG-derived information, inferring required insights and determining question answerability. 
For personalization, explanations adapt to the targeted user level when the persona is provided, indicating information sources more clearly. 
For the general pipeline, the answering model maintains a high-level, neutral tone.

\subsection{Summarization}
\label{sec:extended_result_tables}
We provide extended versions of \Cref{tab:general_pipeline_results,tab:personal_pipeline_results} in \Cref{tab:general_pipeline_results_extended,tab:personal_pipeline_results_extended}, including the standard deviations of the averaged scores and the score deviation for the personalized scores.

\begin{table*}[]
    \centering
    \small
    \begin{tabular}{lccccccc}
    \toprule
    Setup       & INF   & REL   & FAC & OVR\\
    \midrule
    G-infer     & 4.49 ($\sigma$ 0.66) & 4.04 ($\sigma$ 0.51) & 4.78 ($\sigma$ 0.41) & 4.41 ($\sigma$ 0.64) \\
    G-top & 4.33 ($\sigma$ 0.81) & 4.02 ($\sigma$ 0.55) & 4.67 ($\sigma$ 0.52) & 4.24 ($\sigma$ 0.65) \\
    G-all & 4.40 ($\sigma$ 1.17) & 4.11 ($\sigma$ 1.29) & 4.30 ($\sigma$ 1.19) & 4.35 ($\sigma$ 1.01) \\
    \midrule
    G-none   & 4.31 ($\sigma$ 1.39) & 3.70 ($\sigma$ 1.39) & 4.33 ($\sigma$ 1.40) & 3.99 ($\sigma$ 1.34) \\
    GOLD        & 3.79 ($\sigma$ 1.25) & 3.59 ($\sigma$ 1.16) & 4.98 ($\sigma$ 1.90) & 4.12 ($\sigma$ 1.15)\\
    \bottomrule
    \end{tabular}
    \caption{Extended LLM-based Likert scores of the general multi-source meeting summarization pipeline. The deviation is stated in parentheses.}
    \label{tab:general_pipeline_results_extended}
\end{table*}

\begin{table*}[]
    \centering
    \tiny
    \begin{tabular}{lccccc}
    \toprule
    Setup       & INF-P   & REL-P   & FAC   & OVR-P \\
    \midrule
    P-infer+per & 4.25 ($\sigma$ 0.87) | 4.51 ($\sigma$ 0.69) & 3.93 ($\sigma$ 0.72) | 4.06 ($\sigma$ 0.45) & 4.29 ($\sigma$ 0.74) | 4.52 ($\sigma$ 0.93) & 4.56 ($\sigma$ 0.77) | 4.79 ($\sigma$ 0.45)\\
    
    P-per & 4.17 ($\sigma$ 1.01) | 4.57 ($\sigma$ 0.76) & 3.94 ($\sigma$ 0.74) | 4.38 ($\sigma$ 0.60) & 4.33 ($\sigma$ 0.84) | 4.59 ($\sigma$ 0.52) & 4.23 ($\sigma$ 0.99) | 4.50 ($\sigma$ 0.69)\\    
    %     PER & 4.17 ($\sigma$ 1.01) | 4.57 ($\sigma$ 0.76) & 3.94 ($\sigma$ 0.74) | 4.18 ($\sigma$ 0.60) & 4.33 ($\sigma$ 0.84) | 4.59 ($\sigma$ 0.52) & 4.23 ($\sigma$ 0.99) | 4.50 ($\sigma$ 0.69)\\    

    \midrule
    
    P-infer & 4.23 ($\sigma$ 0.86) | 4.52 ($\sigma$ 0.75) & 3.98 ($\sigma$ 0.71) | 4.16 ($\sigma$ 0.66)& 4.72 ($\sigma$ 0.75) | 4.87 ($\sigma$ 0.33) & 4.18 ($\sigma$ 0.60) | 4.35 ($\sigma$ 0.61)\\
    
    P-all & 4.02 ($\sigma$ 1.59) | 4.18 ($\sigma$ 1.36) & 3.97 ($\sigma$ 1.14) | 4.24 ($\sigma$ 0.96) & 4.19 ($\sigma$ 1.32) | 4.38 ($\sigma$ 0.94) & 3.98 ($\sigma$ 1.26) | 4.20 ($\sigma$ 1.10)\\
    \midrule
    P-none & 4.54 ($\sigma$ 1.05) & 4.23 ($\sigma$ 0.96) & 4.68 ($\sigma$ 0.86) & 4.43 ($\sigma$ 0.91) \\
    \bottomrule
    \end{tabular}
    \caption{Extended LLM-based Likert scores of the personalized multi-source meeting summarization pipeline. The deviation is stated in parentheses. As we assess the scores for all four participant roles, we state here the lowest and highest scores achieved.}
    \label{tab:personal_pipeline_results_extended}
\end{table*}

\section{Summaries Examples}
Following, we present model summaries of the first AMI meeting.
The single-source summaries and gold summary are in \Cref{tab:single_source_examples}.
Summaries from the general pipeline are shown in \Cref{tab:general_pipeline_examples}, personalized pipeline summaries are listed in \Cref{tab:personal_pipeline_results}, and summaries stemming from the smaller models are stated in \Cref{sec:practical_sum_examples}.

\begin{table*}[h]
    \centering
    \small
    \begin{tabular}{lp{12cm}}
    \toprule
        \textbf{Setup}       & \textbf{Summary} \\
        \midrule
        Gold &  "The project manager introduced the upcoming project to the team members and then the team members participated in an exercise in which they drew their favorite animal and discussed what they liked about the animal. The project manager talked about the project finances and selling prices. The team then discussed various features to consider in making the remote." \newline \\
        G-none     &  "In the kick-off meeting for a new project, the team, led by project manager Laura, gathered to design an original, trendy, and user-friendly remote control. The team members, including Craig (User Interface), Andrew (Marketing Expert), and David (Industrial Designer), shared their favorite animals to highlight characteristics they value, such as affection, personality, and health, which could inspire the remote's design. The remote is intended to retail for twenty-five Euros, with a production cost cap of twelve fifty Euros, aiming for a fifty percent profit margin. The team discussed the challenges of designing for an international market, considering factors like regional preferences, keypad styles for different languages, and the potential for added functionalities beyond standard remote controls. They also considered the importance of the remote being easy to find and possibly incorporating advanced features like touchscreens. The team agreed on the need for the remote to control multiple devices and possibly offer unique features, like those found in multifunctional devices such as Palm Pilots. The next steps include individual work on functional design, technical functions, and marketing strategies, with the goal of creating a remote that stands out in the market while meeting cost and design commitments." \newline \\
        P-none       &  "The kick-off meeting for the new remote control project was led by Laura, the project manager, with team members Craig (User Interface), Andrew (Marketing Expert), and David (Industrial Designer) in attendance. The project's goal is to design an original, trendy, and user-friendly remote control, with a selling price set at 25 Euros and a production cost cap of 12.50 Euros. The team discussed the importance of the remote being appealing in different international markets, considering factors like regional preferences, keypad styles, and functionality. The idea of a universal remote control was favored, potentially integrating features beyond TV control, inspired by the multifunctionality of devices like palm pilots. The team also considered the design's aesthetic appeal and user experience, such as streamlining the form factor or adding a touchscreen. The next steps involve individual work on functional design, with a focus on technical functions, market requirements, and the actual working design. The project will proceed in three stages, with meetings to follow each stage of individual work."\\
        \bottomrule
    \end{tabular}
    \caption{Example summaries without considering additional sources, generated by GPT4.}
    \label{tab:single_source_examples}
\end{table*}

\subsection{General Pipeline Summaries}
The summary examples of G-infer, G-top, and G-all are displayed in \Cref{tab:general_pipeline_examples}.

\label{sec:general_sum_examples}

\begin{table*}[h]
    \centering
    \small
    \begin{tabular}{lp{12cm}}
    \toprule
    \textbf{Setup}       &  \textbf{Summary} \\
    \midrule
    G-infer     &  "In a kick-off meeting for a new project, a team discusses the design of an original, trendy, and user-friendly remote control. The design process will involve three stages, with individual work and collaborative meetings. The remote control aims to be sold for twenty-five Euros, with a production cost cap of twelve fifty Euros, and will be marketed internationally. The team considers various features, such as combining multiple device controls into one, adding functionalities like controlling house lighting, and possibly incorporating a touchscreen. They also discuss the need for different keypad styles to accommodate various languages and characters, and the importance of the remote being easy to locate if lost. The team acknowledges the challenge of creating a product that stands out in a market where consumers may not consciously assess their purchasing habits for remote controls. The next steps involve the industrial designer working on the actual design, the user interface specialist focusing on technical functions, and the marketing executive considering the product's requirements. The team contemplates whether the remote control will have features beyond controlling the television, aiming for a unique selling point within the set production cost." \newline\\
    G-top       &  "In a project kick-off meeting, the team is tasked with designing a new, original, trendy, and user-friendly remote control. The team members introduce themselves and their roles: Laura is the project manager, Craig is the User Interface Designer, Andrew is the marketing expert, and David is the industrial designer. They discuss their favorite animals to get comfortable with the whiteboard they'll be using for design stages. The team then moves on to project finance, aiming to sell the remote control for twenty-five Euros and not exceed production costs of twelve fifty Euros. They consider the international market, different regions' preferences, and the need for a unique selling point within the production cost limit. The discussion shifts to the design of the remote control. They agree that it should control multiple devices and possibly include extra functionalities like central lighting control, recording notes, and finding the lost remote. They also discuss the importance of a sleek design and the possibility of a touchscreen. The team plans to conduct marketing research, look at existing products, and gather user opinions. They find that many controls are too complicated, with too many buttons and confusing labeling. They prefer simplicity and an attractive look, with advanced functions available but not directly accessible to reduce confusion.The meeting concludes with plans for" \newline \\
    G-all       &  "The Real Reaction project team convened to discuss the design and functionality of a new ergonomic remote control. The prototype revealed insights into the design's ergonomics, particularly the need for articulation to accommodate handedness, primarily at the LED point. The team considered the transferability of function buttons and joystick between right and left-hand use, suggesting a slightly ovoid-shaped joystick for adjustability. The LCD's position was problematic for left-handed use, necessitating articulation for ergonomic retention. Cost considerations were addressed, with a plastic casing and rubber grip proposed for a more biomorphic and ergonomic form. The design aimed for a single curve, essential on the underside for joystick ergonomics. Additional features included a transmitter and speaker for a remote control finder function. The team debated the balance between design features and cost constraints, considering market research, user requirements, and fashion trends in electronics. They discussed the feasibility of various materials, components, and functionalities, such as a bright color for visibility and a special colored case. The evaluation process involved rating the prototype against criteria reflecting the company's goals, marketing strategies, and user requirements. Despite budget limitations leading to compromises in design, such as opting for plastic over rubber and simplifying the LCD, the team aimed to maintain key features like the LCD display and ergonomic form. The meeting concluded with an acknowledgment of the need for further market research and potential for two product versions to test sales impact."\\
    \bottomrule
    \end{tabular}
    \caption{Example summaries generated by GPT4 for the different setups using the general summary pipeline.}
    \label{tab:general_pipeline_examples}
\end{table*}

\subsection{Personalized Summaries}
In \Cref{tab:personal_pipeline_examples} we display summaries from the P-infer+per, P-per, P-infer, and P-all setups on the project manager role.
\Cref{tab:target_reader_examples} shows P-infer-per summaries for the four different target readers.

\label{sec:personal_sum_examples}

\begin{table*}[h]
    \centering
    \small
    \begin{tabular}{lp{12cm}}
    \toprule
        \textbf{Setup}       & \textbf{Summary} \\
        \midrule
        P-infer+per &  "Laura, the Project Manager, led the kick-off meeting for the design of a new remote control, aiming for it to be original, trendy, and user-friendly. The team discussed the project's financial goals, targeting a selling price of 25 Euros and a production cost not exceeding 12.50 Euros. The remote control is intended for international sales, prompting considerations of regional market appeal, keypad styles, symbols, and possibly language characters. The team explored the idea of integrating multiple device controls into one unit, reflecting on the inconvenience of managing several remotes. They also considered adding unique features, such as house lighting control or a find-me function, and pondered whether to include functionalities beyond TV control. The next steps involve individual work on functional design, with the industrial designer focusing on the working design, the user interface specialist on technical functions, and the marketing executive on product requirements. Laura emphasized the importance of balancing creative design with practical constraints like cost and market considerations, ensuring the project meets its brief." \newline \\
        P-per       &  "Laura, as the project manager, led the kick-off meeting for the new remote control design project. The team, including David (Industrial Designer), Craig (User Interface), and Andrew (Marketing Expert), discussed the project's brief to create an original, trendy, and user-friendly remote control with a selling price of 25 Euros and a production cost cap of 12.50 Euros. The team considered the international market's impact on design, with considerations for regional preferences and functionalities. They explored the idea of combining multiple device controls into one remote and adding unique features like lighting control. The importance of a unique selling point was acknowledged, potentially limited by production costs. The next steps involve individual work and subsequent meetings, with the functional design stage up next. Laura will ensure the team receives clear instructions via email for their respective roles." \newline\\
        \midrule
        P-infer     &  "In a kick-off meeting for a new project, the team, led by project manager Laura, discusses the design of an original, trendy, and user-friendly remote control. Craig, the User Interface expert, shares his parents' frustration with multiple remotes, highlighting the need for a device that consolidates control of various household devices. Andrew, the marketing expert, emphasizes the importance of physical affection and personality in his favorite animal, a beagle, which may translate into the remote's design appeal. David, the industrial designer, notes his preference for whales due to his allergy to fur, suggesting a design that accommodates different user needs. The team contemplates international design considerations, such as varying keypad styles for different languages and the potential for unique features beyond traditional TV remote functions, akin to the evolution of palm pilots. They aim to sell the remote for twenty-five Euros, with a production cost cap of twelve fifty Euros, ensuring a fifty percent profit margin. The discussion includes the possibility of adding a touchscreen and streamlining the design to make it more appealing and technologically advanced. The team plans to work on individual tasks and reconvene to discuss the functional design, considering the remote's technical functions and marketing strategy, with the goal of creating a unique" \newline \\
        P-all       &  "The kick-off meeting for the 'Real Reaction' project, led by Project Manager Laura, focused on the design of a new, original, trendy, and user-friendly remote control. The team, including Craig (User Interface), Andrew (Marketing Expert), and David (Industrial Designer), discussed various aspects of the project, such as favorite animal characteristics to inspire design, market considerations for international sales, and the importance of combining multiple device controls into one remote. The team agreed on a selling price of 25 Euros, aiming for a production cost of no more than 12.50 Euros to achieve a 50\% profit margin. They also considered the need for the remote to appeal to different international markets, potentially requiring different keypad styles and symbols. The discussion highlighted the importance of a unique selling point, streamlined design, and additional functionalities like controlling house lighting. The next steps involve individual work on functional design, user interface, and marketing strategies, with a follow-up meeting scheduled to further refine the product."\\
        \bottomrule
    \end{tabular}
    \caption{Example summaries for the different setups using the personalized summary pipeline generated by GPT4.}
    \label{tab:personal_pipeline_examples}
\end{table*}

\begin{table*}[h]
    \centering
    \small
    \begin{tabular}{lp{12cm}}
    \toprule
        \textbf{Target}       & \textbf{Summary} \\
        \midrule
        PM &  "Laura, you as the Project Manager, led the kick-off meeting for the design of a new remote control, aiming for it to be original, trendy, and user-friendly. The team discussed the project's financial goals, targeting a selling price of 25 Euros and a production cost not exceeding 12.50 Euros. The remote control is intended for international sales, prompting considerations of regional market appeal, keypad styles, symbols, and possibly language characters. The team explored the idea of integrating multiple device controls into one unit, reflecting on the inconvenience of managing several remotes. They also considered adding unique features, such as house lighting control or a find-me function, and pondered whether to include functionalities beyond TV control. The next steps involve individual work on functional design, with the industrial designer focusing on the working design, the user interface specialist on technical functions, and the marketing executive on product requirements. Laura emphasized the importance of balancing creative design with practical constraints like cost and market considerations, ensuring the project meets its brief." \newline \\
        UE &  "In the kick-off meeting for a new remote control design project, the team discussed the product's concept, which aims to be an original, trendy, and user-friendly remote control. The remote control is intended to retail at 25 Euros, with a production cost cap of 12.50 Euros. The team considered international market variations, such as keypad styles and symbols, and the potential for different regional pricing strategies. You highlighted the irony of using multiple remotes for integrated systems and suggested exploring additional functionalities, like home lighting control, drawing parallels to the evolution of palm pilots. They also emphasized the importance of a practical and streamlined design, possibly incorporating a touchscreen. The team debated whether to include features beyond TV control and considered the unique selling point of the product. The next steps involve individual work on functional design, with the User Experience role focusing on technical functions and ensuring the remote control meets user requirements. The production cost was identified as a key factor influencing the design's features and capabilities." \newline \\
        ID &  "David, as the Industrial Designer, your focus in the kick-off meeting was on creating a remote control that is original, trendy, and user-friendly, with a strict production cost limit of 12.50 Euros to maintain a retail price of 25 Euros. You emphasized the importance of universal functionality to reduce the number of remotes needed, drawing from Craig's anecdote about his parents' frustration with multiple remotes. You also considered the potential for adding new features, akin to the evolution of palm pilots, and suggested the remote could control additional household devices, like lighting.You highlighted the need for international compatibility, taking into account different keypad styles and symbols for various languages, which could impact the design and production cost. The discussion about the remote's appeal in different international markets indicated that pricing and features might need to be tailored to regional preferences.You were concerned about the remote being easily lost and suggested incorporating a find-me feature. A touch screen was proposed for a modern look, aligning with the trend of chic home devices. You questioned whether the remote should include features beyond TV control, indicating a need for clarity on the project's scope. For the next steps, you will work on the actual design, balancing the desired features with the production cost" \newline \\
        M  &  "Andrew, as the marketing expert, your focus during the kick-off meeting for the new remote control project was on ensuring the product's marketability. You emphasized the importance of understanding regional market differences, suggesting that the €25 selling price might vary in appeal across different areas. You also highlighted the need for the remote to stand out with unique features, considering the addition of functionalities beyond standard TV control, such as home lighting management, to create a unique selling point. You questioned whether the design should be limited to TV features or include broader capabilities, reflecting on the success of multifunctional devices like Palm Pilots. Additionally, you were concerned with the user experience, noting the common issue of losing remotes and suggesting a streamlined design, possibly with a touchscreen. The meeting covered the project's financial goals, aiming for a production cost cap at 50\% of the selling price, and the need for the design to be original, trendy, and user-friendly. The next steps involve individual work and further meetings to refine the functional design, with you focusing on the product's market requirements." \\
        \bottomrule
    \end{tabular}
    \caption{Example summaries for different participants, showing the individual adaptions. Summaries are generated using the personalization pipeline with GPT4 as backbone model.}
    \label{tab:target_reader_examples}
\end{table*}

\subsection{Practical Setup Summaries}
\label{sec:practical_sum_examples}
In \Cref{tab:practical_general_pipeline_examples,tab:practical_personal_pipeline_examples} we show the summaries of the smaller models Gemini, Phi3, Llama3 on the first AMI meeting with the general G-infer and the personalized P-infer+per setups, respectively.

% GENERAL
\begin{table*}[h]
    \centering
    \small
    \begin{tabular}{lp{12cm}}
    \toprule
        \textbf{Target}       & \textbf{Summary} \\
        \midrule
        Gemini & "This transcript captures a kick-off meeting for a project to design a new remote control. The team, consisting of a project manager, an industrial designer, a user interface expert, and a marketing expert, are tasked with creating a remote control that is original, trendy, and user-friendly. The design process will involve three stages of individual work followed by group discussions. The team discusses the target market, the desired features, and the potential challenges of designing a remote control that is both functional and appealing. They also consider the possibility of incorporating features beyond basic television control, such as controlling other devices in the home. The meeting concludes with a discussion of the next steps, which include developing the functional design of the remote control." \newline \\
        Phi3 &  "The project kickoff meeting introduces the design of a new, trendy, and user-friendly remote control. The team will engage in individual work and iterative meetings, focusing on user interface, technical functions, and marketing requirements. They consider the remote's price point of 25 Euros, aiming for a production cost of no more than 12.5 Euros and a 50\% profit margin. International market considerations, such as varying economic conditions and technological advancements, will influence design decisions. The team discusses potential unique features, like combining multiple functions, adding a touch screen, and creating a design that minimizes loss. They also consider the remote's appeal in different markets and the importance of a cost-effective yet premium design." \newline \\
        Llama3  &  "The project manager introduced the upcoming project to the team members and then the team members participated in an exercise in which they drew their favorite animal and discussed what they liked about the animal. The project manager talked about the project finances and selling prices. The team then discussed various features to consider in making the remote." \\
        \bottomrule
    \end{tabular}
    \caption{General summary examples generated by three practical models, Gemini, Llama3, Phi3.}
    \label{tab:practical_general_pipeline_examples}
\end{table*}

% PERSONAL
\begin{table*}[h]
    \centering
    \small
    \begin{tabular}{lp{12cm}}
    \toprule
        \textbf{Target}       & \textbf{Summary} \\
        \midrule
        Gemini &  "The meeting was a kick-off for the design of a new remote control. The goal is to create a product that is original, trendy, and user-friendly. The design process will involve three stages: individual work, a meeting to discuss the work, and then repeating the process three times. The remote control will be sold for 25 Euros and should cost no more than 12.50 Euros to produce. The team discussed the potential impact of international markets on the design, including the possibility of different keypad styles and symbols for different regions. They also considered the possibility of adding features beyond television control, such as controlling lighting or integrating with other devices like palm pilots. The team will continue to work on the design in the next stage, focusing on functional design."  \newline \\
        Phi3 &  "The project team, led by Project Manager Laura, is tasked with designing a new, original, trendy, and user-friendly remote control, with a selling price of 25 Euros and a production cost not exceeding 12.5 Euros. The design will consider international sales, potentially incorporating features like multiple zones, characters, and keypad styles to cater to various languages and regions. The team also discusses the need for a multi-functional design, possibly integrating with other devices, and a user-friendly interface with innovative features like a touch screen or voice command. The design process will involve iterative stages of individual work and meetings, with the industrial designer focusing on technical aspects and the marketing executive on requirements and features. The team also acknowledges the common issue of losing remote controls and considers designing a more integrated or innovative solution to address this problem." \newline \\
        Llama3  &  "The meeting discusses the design and features of a remote control. Craig, the User Interface expert, shares his favorite animal, the monkey, and his parents' experience with multiple remote controls. Andrew, the marketing expert, shares his favorite animal, the beagle, and its characteristics, such as being willing to take physical affection and having lots of personality. The discussion highlights the importance of considering production cost, selling price, and international pricing when designing the remote control. The team also explores the idea of combining multiple functions into one remote control, citing the example of palm pilots, which evolved from simple devices to multifunctional tools. The meeting concludes with a discussion on how to make the remote control more streamlined and appealing to users, possibly by incorporating new features like lighting control." \\
        \bottomrule
    \end{tabular}
    \caption{Personalized summary examples generated by three practical models, Gemini, Llama3, Phi3.}
    \label{tab:practical_personal_pipeline_examples}
\end{table*}

\section{Prompt Templates}
\label{sec:prompt_templates_appendix}
In the following, we present the prompt templates used to identify gaps in a given transcript (\Cref{fig:gap_prompt}), infer information from a set of related documents (\Cref{fig:inferring_prompt}), summarize the enriched transcript (\Cref{fig:summarization_prompt}) and extract a persona (\Cref{fig:persona_prompt}).
The persona related prompt-passages are optional and left out for the general summarization pipeline.

\begin{figure*}[t]
    \begin{AIbox}{Gap Identification Prompt Template}
    \parbox[t]{\textwidth}{
    For the following task, respond in a way that matches this description: <persona>. \newline
    Take the role of a question generator that takes the role of a defined participant and points out unclarities and open questions in a transcript.
    Generate at most 5 questions. Only ask the 5 most relevant questions. \newline
    \newline
    If you were participant <participant>, what open questions would you still have in regards to the following transcript: <transcript>? \newline
    \newline
    Your answer shall only contain a Python array of dictionaries: '[{<question>, <insert>}, {<question>, <insert>}, {<question>, <insert>}, ...]'.
    Each dict must contain an entry called 'question' containing the question itself and an entry called 'insert' containing an exact copy of the sentence from the transcript that is most relevant to the question.
    }
    \end{AIbox}
    \caption{Gap identification prompt template.}
    \label{fig:gap_prompt}
\end{figure*}

\begin{figure*}[t]
    \begin{AIbox}{Salient Information Inferring Prompt Template}
    \parbox[t]{\textwidth}{
    Format your entire answer as a JSON object, with an entry named "answer" containing your answer and an entry "able" containing a binary value (true or false, all lower case) for whether you were actually able to answer the question.\newline
    Base your answer strictly on information contained in the prompt, without speculating.
    Tailor your answer so it fits best to this persona: <persona>.\newline
    The answer should be a single running text string, not a list or dictionary. \newline
    \newline
    Answer based on the following transcript and a supplemental file. \newline
    Transcript: <transcript> \newline
    Supplemental file: <file> \newline
    }
    \end{AIbox}
    \caption{Information inferring prompt template.}
    \label{fig:inferring_prompt}
\end{figure*}

\begin{figure*}[t]
    \begin{AIbox}{Abstractive Summarization Prompt Template}
    \parbox[t]{\textwidth}{
    You are a professional summarizer and have been tasked with creating an abstractive summary for a participant in a meeting.
    Your summary should be 250 tokens or less.
    Carefully analyze the following transcript and provide a detailed summary for the participant.
    Consider the target persona who will have to work with the summary: <{persona}>.\newline
    The generated summary should help the persona understand the meeting content even after a long time, and it should be the perfect source for the persona to post-process the meeting content and prepare for the next steps.
    Focus on what is relevant for the participant to know and add what the participant needs to know to best work with the meeting content. \newline
    \newline
    Summarize this transcript. Create an abstractive summary. Make the summary 250 tokens or less. \newline
    Transcript: <enriched transcript>
    }
    \end{AIbox}
    \caption{Abstractive summarization prompt template for enriched transcript.}
    \label{fig:summarization_prompt}
\end{figure*}

\begin{figure*}[t]
    \begin{AIbox}{Persona Extraction Prompt Template}
    \parbox[t]{\textwidth}{
    You are a professional profiler and have been tasked with creating a persona for a participant in a meeting.
    Carefully analyze the following transcript and provide a detailed persona for the participant.
    In your answer, include the participant's role, personality traits from the Big Five, point of view, contributions, knowledge that they brought to the meeting, information that they did not know, and any other relevant information.
    Make sure to provide a detailed and comprehensive persona.
    Your answer should be a string containing a running text. \newline
    \newline
    Create a persona for participant <participant> based on the following transcript: <transcript>.

    }
    \end{AIbox}
    \caption{Persona extraction prompt template.}
    \label{fig:persona_prompt}
\end{figure*}

\begin{figure*}[t]
    \begin{AIbox}{Evaluation Prompt Template}
    \parbox[t]{\textwidth}{
        You are an expert in the field of summarizing meetings and are tasked with evaluating the quality of the following summary.
        Score the summary according to the scoring criteria with a Likert score between 1 (worst) and 5 (best). \newline
       \newline
        Transcript: <{transcript}>\newline
        Summary: <{summary}>\newline
        Criteria: <{criteria}> \newline
        \newline
        Your task is to rank the summaries based on the criteria provided.
        Remember to consider the quality of the summaries and how well they capture the key points of the original transcript.
        First provide an argumentation for your ranking. Therefore, use chain-of-thought and think step by step.
        Return a json object with the ranking for the evaluation criteria.
        The output should be in the following format:
        <explanation, step-by-step> ! <json object>
        The json object should follow the structure ```json {<evaluation criteria> : <Likert Score>}```
        The JSON object should only contain the single Likert score for the currently assessed criteria.
    }
    \end{AIbox}
    \caption{All-in-one evaluation prompt template.}
    \label{fig:evaluation_prompt}
\end{figure*}

\clearpage
\onecolumn
\hypertarget{annotation}{}
\citationtitle

\begin{bibtexannotation}
@inproceedings{kirstein-etal-2024-multisource,
 author={Frederic Kirstein, Terry Ruas, Robert Kratel, Bela Gipp},
 title={Tell me what I need to know: Exploring LLM-based (Personalized) Abstractive Multi-Source Meeting Summarization},
 booktitle={Proceedings of the 2024 Conference on Empirical Methods in Natural Language Processing: Industry Track},
 pages={20},
 publisher={Association for Computational Linguistics},
 year={2024},
 month={11}
}\end{bibtexannotation}

\end{document}